\documentclass[conference]{IEEEtran}
\IEEEoverridecommandlockouts

\usepackage{cite}
\usepackage{amsmath,amssymb,amsfonts}
\usepackage{graphicx}
\usepackage{algorithmic}
\usepackage{graphicx}
\usepackage{textcomp}
\usepackage{xcolor}
\usepackage{booktabs}
\usepackage{tcolorbox}
\usepackage{hyperref}
\usepackage{bbold}
\usepackage{balance}
\def\BibTeX{{\rm B\kern-.05em{\sc i\kern-.025em b}\kern-.08em
    T\kern-.1667em\lower.7ex\hbox{E}\kern-.125emX}}

\begin{document}

\title{Can Agentic AI Match the Performance of\\ Human Data Scientists?
\thanks{This paper is based upon work supported by the Cisco Research gift fund and National Science Foundation under CAREER Grant No. 2338506.}
}
\author{
An Luo\IEEEauthorrefmark{1},
Jin Du\IEEEauthorrefmark{1},
Fangqiao Tian\IEEEauthorrefmark{1},
Xun Xian\IEEEauthorrefmark{2},
Robert Specht\IEEEauthorrefmark{1},
Ganghua Wang\IEEEauthorrefmark{3},\\
Xuan Bi\IEEEauthorrefmark{4},
Charles Fleming\IEEEauthorrefmark{5},
Jayanth Srinivasa\IEEEauthorrefmark{5},
Ashish Kundu\IEEEauthorrefmark{5},
Mingyi Hong\IEEEauthorrefmark{2},
Jie Ding\IEEEauthorrefmark{1}
\\[1ex]
\IEEEauthorrefmark{1}School of Statistics, University of Minnesota, Minneapolis, MN, USA\\
\IEEEauthorrefmark{2}Department of Electrical and Computer Engineering, University of Minnesota, Minneapolis, MN, USA\\
\IEEEauthorrefmark{3}Data Science Institute, University of Chicago, Chicago, IL, USA\\
\IEEEauthorrefmark{4}Carlson School of Management, University of Minnesota, Minneapolis, MN, USA\\
\IEEEauthorrefmark{5}Cisco Research, San Jose, CA, USA\\
}

\maketitle

\begin{abstract}
Data science plays a critical role in transforming complex data into actionable insights across numerous domains. Recent developments in large language models (LLMs) have significantly automated data science workflows, but a fundamental question persists: Can these agentic AI systems truly match the performance of human data scientists who routinely leverage domain-specific knowledge? We explore this question by designing a prediction task where a crucial latent variable is hidden in relevant image data instead of tabular features. As a result, agentic AI that generates generic codes for modeling tabular data cannot perform well, while human experts could identify the important hidden variable using domain knowledge. We demonstrate this idea with a synthetic dataset for property insurance. Our experiments show that agentic AI that relies on generic analytics workflow falls short of methods that use domain-specific insights. This highlights a key limitation of the current agentic AI for data science and underscores the need for future research to develop agentic AI systems that can better recognize and incorporate domain knowledge.
\end{abstract}


\section{Introduction}

Data science is a central interdisciplinary field that blends statistics, computer science, and domain expertise to extract actionable insights from complex, heterogeneous data~\cite{Cao2017DataS, brodie2023axiology}. By transforming raw information into knowledge and value, data science drives innovation and shapes decision-making in science, industry, healthcare, finance, and beyond~\cite{Grossi2021DataSA,Blair2019DataSO}.

Recent advancements in large language models (LLMs) have significantly accelerated the automation of data science workflows. LLMs such as GPT-4~\cite{achiam2023gpt} and Claude~\cite{anthropic2025claude3.7} have demonstrated impressive capabilities in automating code generation and executing regular machine learning tasks~\cite{Jiang2025AIDEAE, Hong2024DataIA,  Liang2025IMCTSEA,Li2024AutoKaggleAM}. These developments offer promising potential for streamlining common analytical processes and reducing the manual workload of human data scientists.

Despite these advancements, there is still a lack of understanding of whether agentic AI really perform as good as human data scientists. In practice, human data scientists consistently rely on specialized knowledge about the data or task and incorporate crucial nuances that enhance model performance~\cite{Mao2019HowDS, Zhang2020HowDD, Lin2025spike, Lin2025spatial, Luo2025AssistedDSBH}. Such domain-driven decisions are often subtle yet essential, as they address complexities not captured by typical analytics workflows. 
However, current research on LLM-driven data science has largely focused on generating generic code and pipeline executions~\cite{Li2024AutoKaggleAM, Jiang2025AIDEAE}. These approaches often neglect the domain-specific knowledge needed for complex, real-world problems. Meanwhile, existing evaluation benchmarks such as MLE-bench~\cite{chan2025mlebench} and DSBench~\cite{jing2025dsbench} aim to assess predictive performance, but do not test whether agentic AI can effectively leverage domain insights outside tabular data. The above observations motivate a fundamental question: \emph{Can agentic AI, which typically relies on generic code generation, truly match the performance of human data scientists who could apply domain knowledge?}

\begin{figure}[htbp]
\centering
\includegraphics[width=0.45\textwidth]{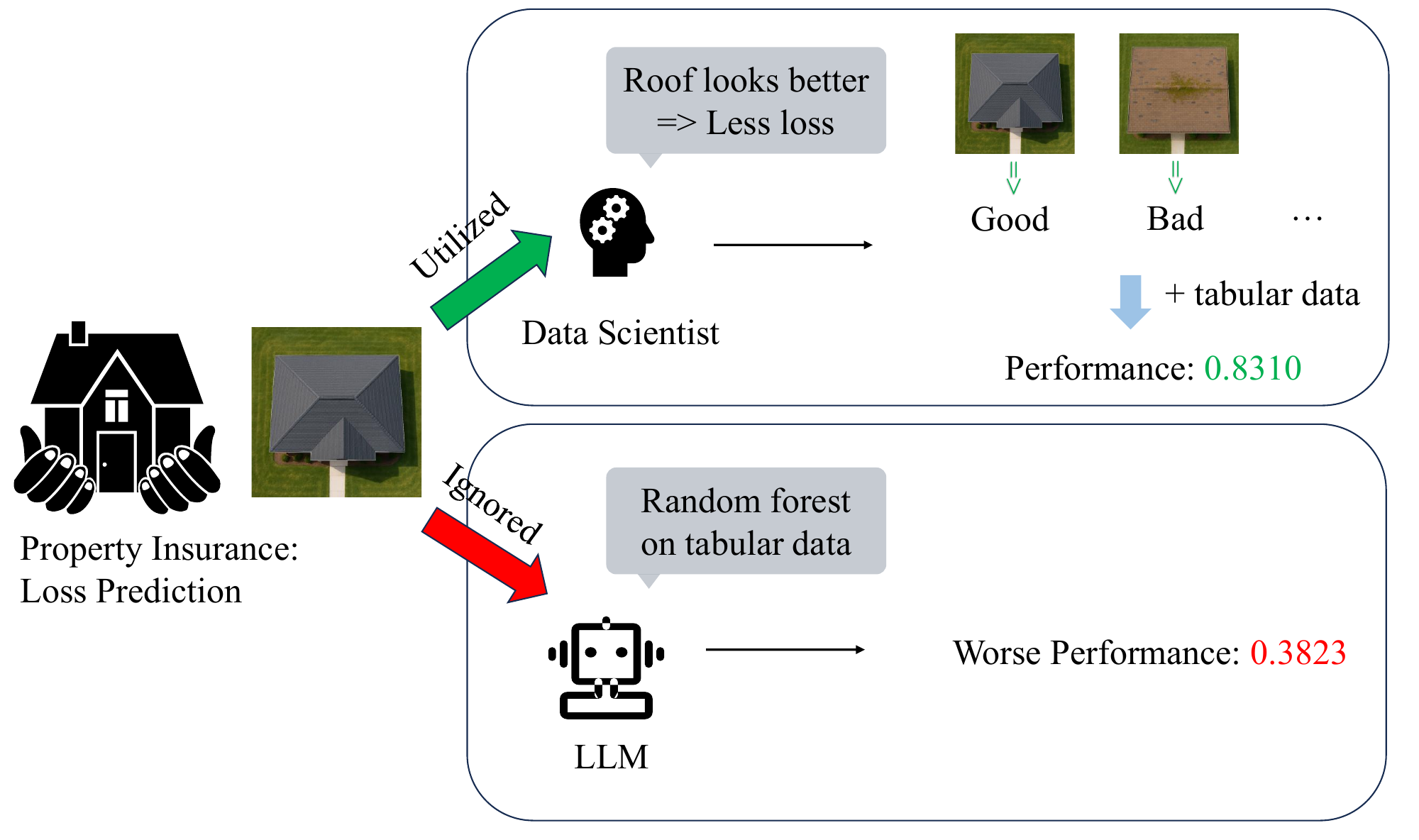}
\caption{\textbf{Comparison of data scientists and agentic AI approaches to loss prediction in property insurance.} The upper panel shows how a human data scientist leverages domain knowledge: by interpreting roof images to infer the critical latent variable (Roof Health) and incorporating it with tabular data, they can achieve substantially high predictive performance (normalized Gini  $=0.8310$). The lower panel depicts an agentic AI's approach, which applies standard tabular modeling while ignoring the image modality and domain-specific cues, resulting in much worse performance (normalized Gini $= 0.3823$). This demonstrates the empirical gap between human data scientist and agentic AI performance when domain knowledge is necessary.}
\label{fig:abstract}
\end{figure}

To address this question, our paper presents an experimental study on a carefully curated dataset that mimics the complexity of real-world data science problems.
The use of synthetic data allows us to control key latent variables that introduce complexity behind observed feature variables. This controlled setup reveals important differences: human data scientists are often able to explicitly identify and leverage domain-specific cues, whereas agentic AI may rely on generic algorithms that do not fully capture the influence of latent factors. Figure~\ref{fig:abstract} illustrates this idea in our property insurance setup.

Our main contributions are summarized below.
\begin{itemize}
\item We design a synthetic dataset that clearly illustrates a fundamental gap between agentic AI (which generates generic, tabular-focused code) and human data scientists (who leverage domain knowledge embedded in images).
\item We empirically quantify this performance gap and demonstrate the importance of domain knowledge in achieving excellent prediction performance.
\end{itemize}
Through this work, we aim to highlight the need for future research to improve agentic AI's ability to identify and use domain-specific knowledge from multimodal data sources.

\section{Design of Synthetic Data}\label{sec:pricing}

\subsection{General Design Principles}

To evaluate the limits of both human data scientists and agentic AI, we generate a controlled synthetic dataset with a hidden latent factor that affects the prediction target but is not present in the tabular features. Instead, this latent variable is embedded in a secondary modality (here, overhead images) to ensure it can be accessed only through domain knowledge and not by generic code.
While the approach could generalize to other modalities such as text or audio, we focus on images to illustrate the mechanism. 
Importantly, these images are crafted so that a knowledgeable data scientist can interpret the latent variable in the context of property insurance, making the challenge meaningful and realistic.

To accomplish this, we use a text-to-image model with engineered prompts to ensure that the generated images faithfully reflect the intended values of the latent variable. This design allows us to examine the gap between generic AI pipelines and human data scientists that can look for the incorporation of domain knowledge.

\subsection{Data Curation}

The data science task specifies a policy table as tabular dataset, and each policy is associated with an image.
Their goal is to predict each home's total insured loss in the next policy
year, \(Y_p\).  The key latent variable is 
\emph{RoofHealth}, a three-level variable—\texttt{Good}, \texttt{Fair}, or \texttt{Bad}. This variable is never shown in the policy table but can be inferred from the image. Below we present how we create this synthetic dataset.

\subsection*{Step 1: Generate Structured Policy Features}
For each policy \(p=1,\dots,n\) we draw:
\begin{enumerate}
  \item PolicyID: ``POL-000001'', \dots
  \item HouseValue: \(X_{\mathrm{val},p}\sim\mathrm{LogNormal}(12.9,\,0.45)\)
       (median \(\approx\$403\mathrm{k}\)).
  \item HouseAge:   \(X_{\mathrm{age},p} \sim 120\,\mathrm{Beta}(4,3)\)
        \;(\(\approx\!40 \) yr median).
  \item WallType:   \(X_{\mathrm{wall},p}\sim \textrm{Bernoulli}(\text{Wood},\text{Brick})\)
        with probabilities \(\{0.9,0.1\}\).
  \item AreaRisk:   \(X_{\mathrm{risk},p}\sim\mathrm{Beta}(2,5)\)
        (0–1 storm exposure).
  \item CreditScore: \(X_{\mathrm{cred},p}\) drawn from the US
        FICO distribution (300–850).
  \item \textbf{RoofHealth} (latent): compute
\[
\begin{split}
  S_p &= 0.02\,X_{\mathrm{age},p}
         +3.0\,X_{\mathrm{risk},p}
         -2.0\bigl(X_{\mathrm{cred},p}/850\bigr)\\
      &\quad+\varepsilon_p,\quad
         \varepsilon_p\sim N(0,1)\,,
\end{split}
\]
then assign \texttt{Good}, \texttt{Fair}, or \texttt{Bad} by partitioning \(S_p\) at the 55th and 80th percentiles of all scores.
\end{enumerate}
Only columns 1–6 are released as features in the policy table.

\subsection*{Step 2: Create Roof Images}
\begin{figure}[htbp]
  \centering
  \begin{minipage}[b]{0.15\textwidth}
    \includegraphics[width=\linewidth]{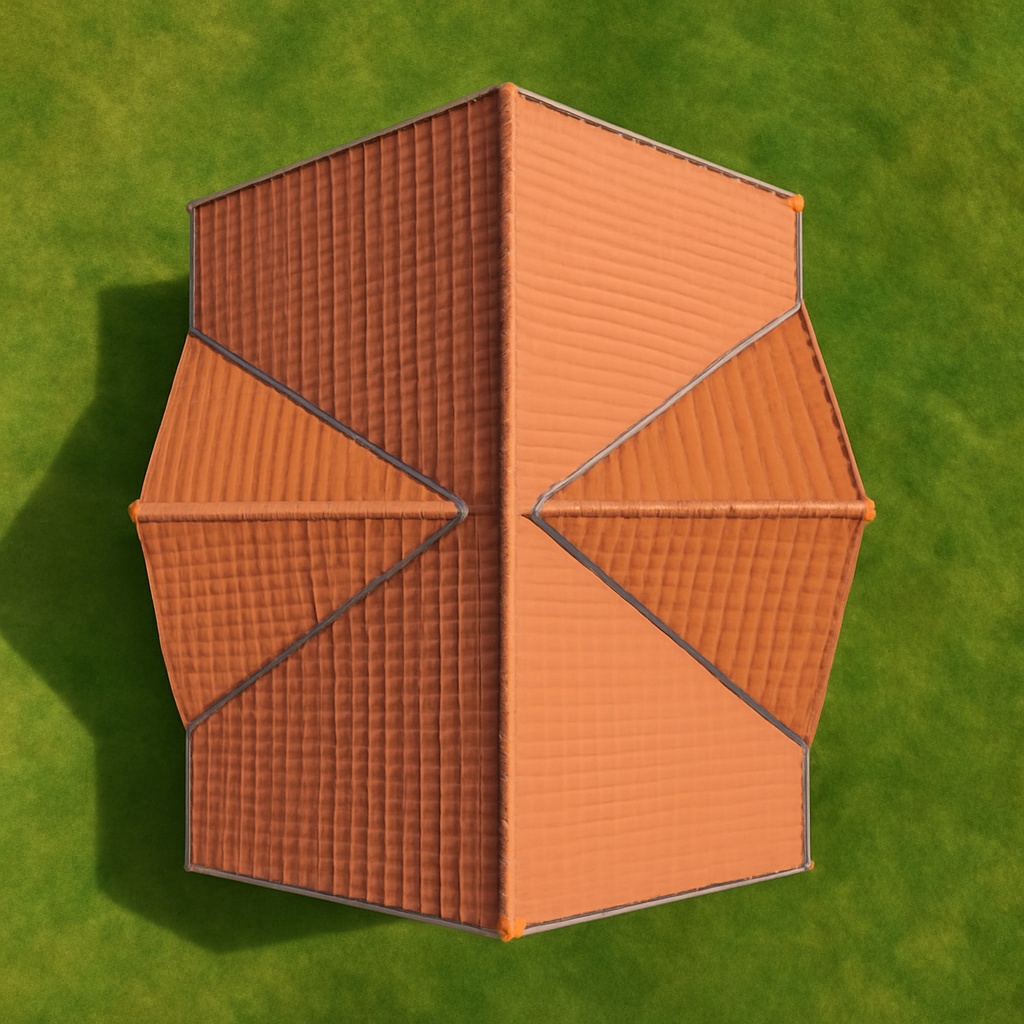}
  \end{minipage}
  \hfill
  \begin{minipage}[b]{0.15\textwidth}
    \includegraphics[width=\linewidth]{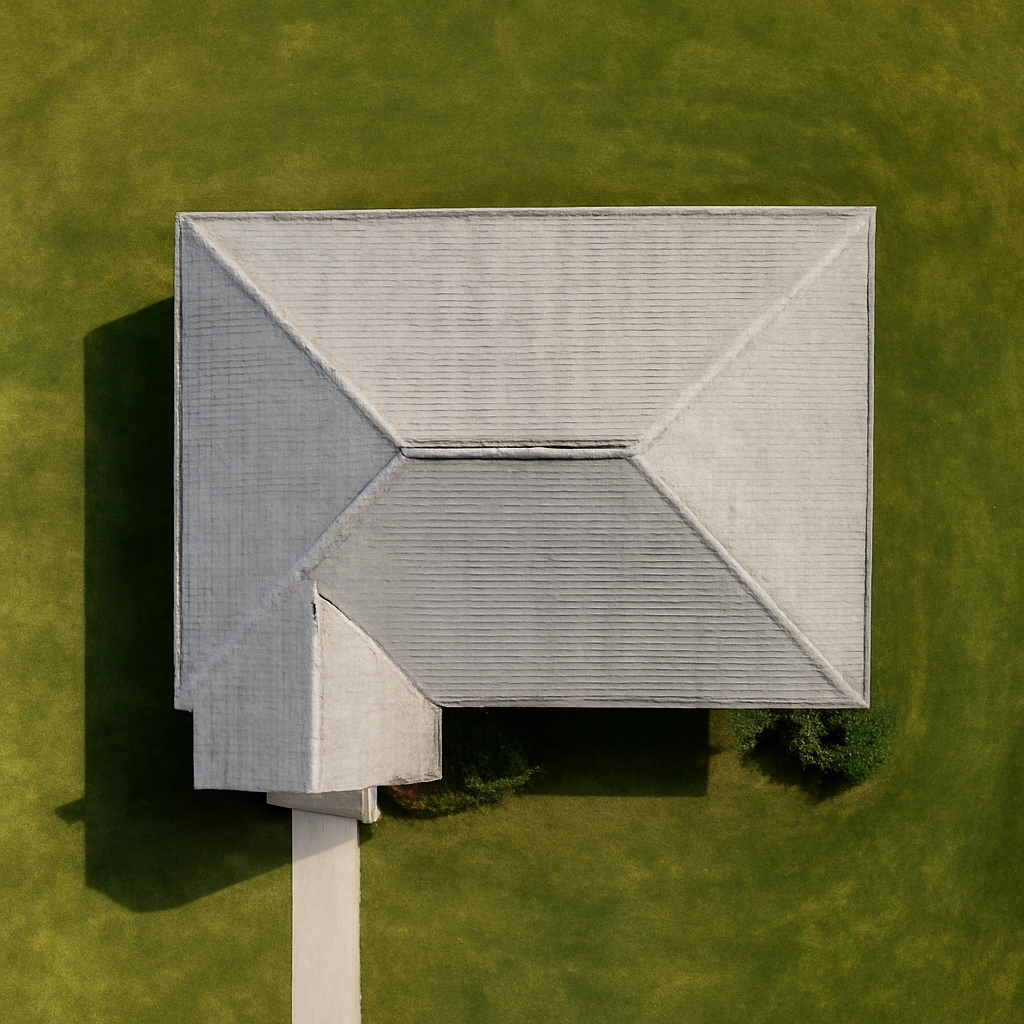}
  \end{minipage}
  \hfill
  \begin{minipage}[b]{0.15\textwidth}
    \includegraphics[width=\linewidth]{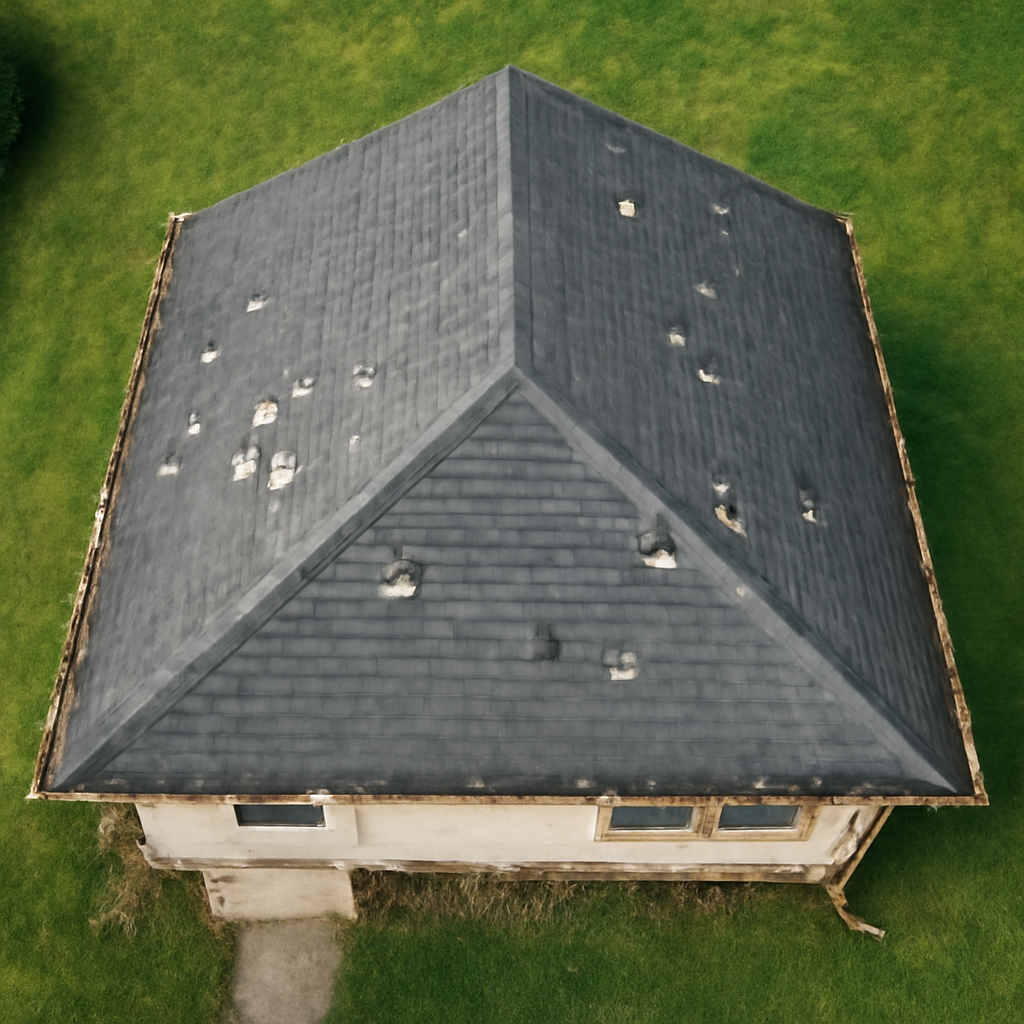}
  \end{minipage}
  \caption{Example overhead roof images generated for our synthetic property insurance dataset. Each image visually encodes a key latent variable, \texttt{RoofHealth}, with three possible states: (a) Good, (b) Fair, and (c) Bad. This variable is never released directly in the tabular data but can be inferred from domain-specific visual cues, such as surface, edge, and extra details such as flashing condition and debris. To create this setting, we use a text-to-image model with carefully designed prompts to ensure each image faithfully represents the intended roof condition. This design allows us to rigorously compare standard agentic AI pipelines (which use only tabular data) against approaches or human experts capable of incorporating additional domain knowledge from the image modality.}\label{fig:vc}
\end{figure}

Each policy gets one 1024\,$\times$\,1024 PNG image  
 synthesised with \texttt{gpt-image-1} using the prompt template below:
\begin{tcolorbox}[colback=gray!05,colframe=gray!60,title=Prompt Template For Roof Image Generation]
\small\texttt{
Realistic straight-down aerial photo of a detached house,
full roof and surrounding lawn in view,
\{\textit{roof\_style}\} roof with \{\textit{shingle\_color}\} shingles,
\{\textit{surface descriptor}\}, \{\textit{edge descriptor}\},
\{\textit{extra descriptor}\}.}
\end{tcolorbox}
In the template, the roof style is sampled from \{\textit{gable, hip, flat, mansard, shed}\};
shingle color is from \{\textit{dark-gray, light-gray, brown, black, red-tile}\}.
Descriptors are from examples as shown in Table \ref{tab:prompt-parts} so that
\texttt{Good/Fair/Bad} roofs differ in surface integrity, edge
condition, and so on. In this way, each image faithfully represents the intended roof condition. See examples of generated images in Figure~\ref{fig:vc}.

\begin{table}[htbp]
\centering
\footnotesize
\caption{Descriptor examples used in prompt template for images generation under each Roof Health category}
\label{tab:prompt-parts}
\begin{tabular}{@{}lcc@{}}
\toprule
\textbf{Roof Health} & \textbf{Surface Example} & \textbf{Edge Example} \\ \midrule
Good & even rows of intact shingles  & well-sealed ridge lines \\
Fair & slightly faded shingles       & ridge line with mild wear \\
Bad  & multiple missing shingles     & damaged or sagging ridge \\
\bottomrule
\end{tabular}
\end{table}

\subsection*{Step 3: Simulate Next-Year Loss \(Y_p\)}
\begin{figure}
    \centering
    \includegraphics[width=1.0\linewidth]{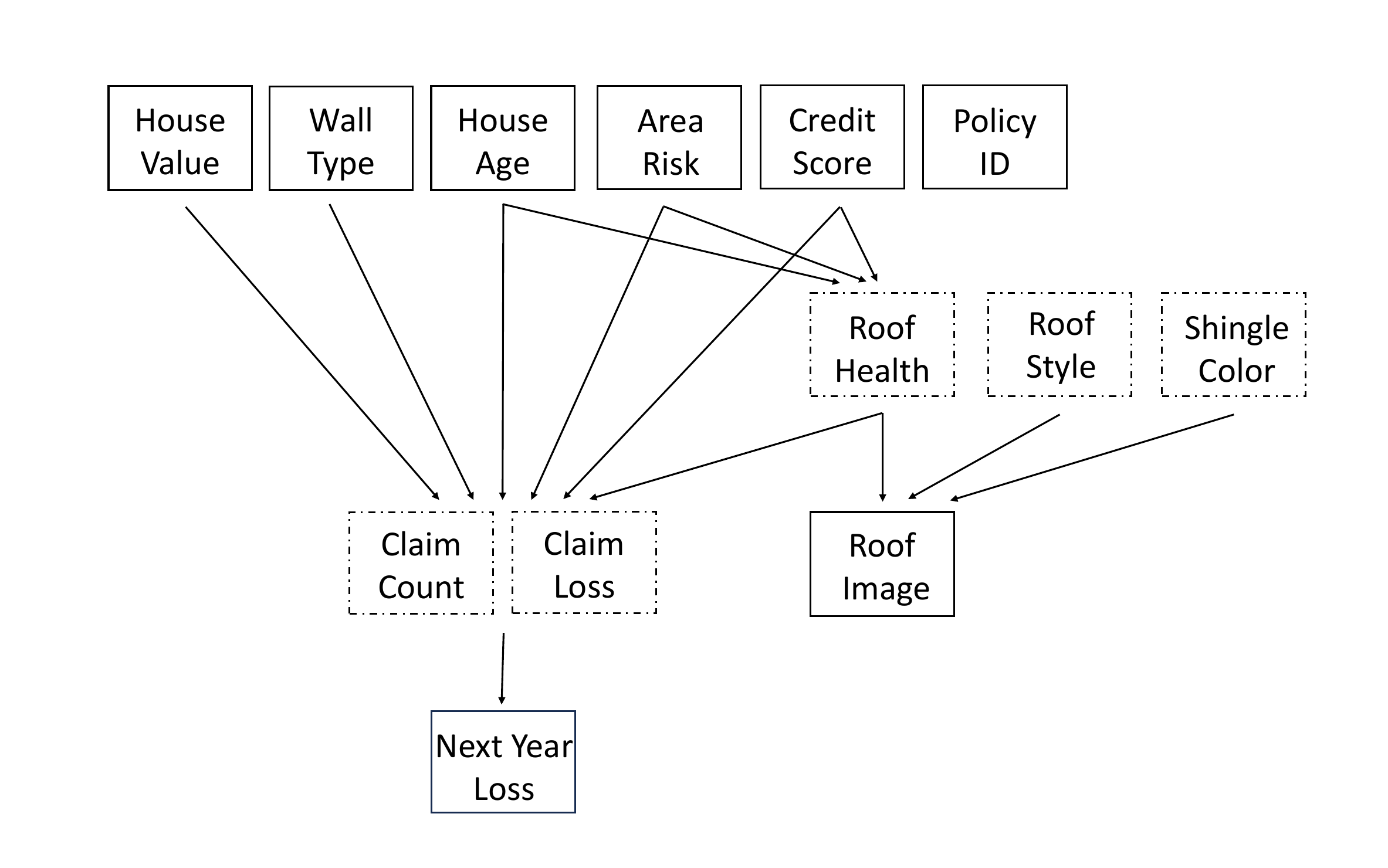}
    \caption{Illustration of data generating process for property insurance. The diagram shows how each policy’s outcome, Next-Year Loss, is generated. Dotted lines surround latent variables that are hidden. The process unfolds as follows: (1) Structured policy features (e.g., House Value, House Age, Wall Type, Area Risk, Credit Score) are generated for each policy. (2) A latent variable, \texttt{Roof Health} (Good, Fair, Bad), is determined by a function of selected features, but is not included in the released tabular data. Instead, it is visually encoded in an accompanying roof image, which is generated for each policy using a random combination of roof style and shingle color. (3) Claim count and claim loss are simulated using both policy features and the latent RoofHealth. The total insured loss for the next year ($Y_p$) is calculated as the sum of all claim loss. To achieve optimal prediction, the hidden \texttt{Roof Health} must be inferred from the roof image.}
    \label{fig:challenge_2}
\end{figure}

\paragraph{Claim count}
With \(\alpha_{\mathrm{rh}}=\{0,1.2,2.4\}\) for \texttt{Good/Fair/Bad},  
\begin{equation}\label{eq:frequency}
\begin{split}
  \lambda_p &= \exp\Bigl(
      -3.0
      +0.03\,\ln\!\frac{X_{\mathrm{val},p}}{250{,}000}
      +0.01\,X_{\mathrm{age},p}\\
      &\quad\;+\;0.05\,X_{\mathrm{risk},p}
      +\alpha_{\mathrm{rh}}\!\bigl(\text{RoofHealth}_p\bigr)
  \Bigr)\,,\\
  N_p &\sim \mathrm{NegBinom}\bigl(r=10,\;\text{mean}=\lambda_p\bigr)\,.
\end{split}
\end{equation}
\paragraph{Claim loss}
With \(\beta_{\mathrm{rh}}=\{0,1.0,2.0\}\),  
\begin{equation}\label{eq:severity}
\begin{split}
  \mu_p &= 7.0
        +0.02\,\mathbb{1}\bigl(X_{\mathrm{wall},p}=\mathrm{Wood}\bigr)
        +0.02\,X_{\mathrm{risk},p}\\
        &\quad\;+\;\beta_{\mathrm{rh}}\!\bigl(\text{RoofHealth}_p\bigr)\,,\\
  Z_{p,j} &\sim \Gamma\!\bigl(k=2,\;\theta=\exp(\mu_p)/2\bigr)\,.
\end{split}
\end{equation}
\paragraph{Total loss}
\[
  Y_p = \sum_{j=1}^{N_p} Z_{p,j}, \qquad
  \text{if }N_p=0 \text{ then } Y_p=0.
\]
The training file exposes \texttt{NextYearLoss} \(=Y_p\); the test file
omits it for evaluation. See Figure~\ref{fig:challenge_2} for an illustration of how the outcome variable, Next-Year Loss, is generated.

The construction of our synthetic property insurance dataset is grounded in established actuarial practice and empirical research. Roof condition is an important factor in property risk and claims, but it is often not directly available in tabular data~\cite{Alzarrad2022AutomaticAO,EvaluatingHailDamageUsingPropertyInsuranceClaimsData}. Our use of roof images is intended to reflect this real-world limitation. The target outcome, next-year loss, is generated using a compound frequency-severity model. This approach is a standard actuarial method for property insurance loss modeling~\cite{GARRIDO2016205,Frees_Derrig_Meyers_2014a}. Together, these design choices ensure our dataset realistically captures the complexities of property insurance prediction.

\section{Performance Study: Generic Pipeline from AI vs. Domain Knowledge Usage from Human}

\subsection{Experimental Setting}

Our goal is to evaluate how the use of domain knowledge impacts predictive performance when crucial information is embedded in the roof images. To do this, we compare three groups of modeling strategies. The first group simulates a generic agentic AI approach that uses only tabular data. The second group represents methods a human data scientist might use, combining tabular data with different ways of extracting information from roof images. The final group is an oracle model that has access to the true latent variables and the underlying data generation process, serving as the best achievable benchmark. By comparing these approaches, we can quantify the value of domain knowledge and highlight the limitations of generic AI workflows.

Below, we outline the specific modeling strategies in each group and how they reflect use of domain knowledge:
\begin{enumerate}
\item \textbf{Agentic AI (Generic pipeline)} Only the tabular features are used and no image data is included. This matches how agentic AI would typically apply generic code to a standard tabular prediction problem. 
\item \textbf{Data Scientists (Image use).}  
Both tabular features and image data are utilized. We consider several practical ways a human data scientist might incorporate image information. One approach is to extract features from the images using a pretrained CLIP model~\cite{Radford2021LearningTV} and either use these features directly or cluster them into categories for use in the predictive model. Another approach is to apply a vision-language model \texttt{gpt-4o-mini} to extract the RoofHealth label from the images. Finally, we include an ideal scenario where the data scientist perfectly labels the true RoofHealth for each image. This represents the best possible use of domain expertise.
\item \textbf{Oracle (Best achievable).} This method uses the exact data-generation formulas and the true \texttt{RoofHealth}. It calculates the predicted loss as the exact product of expected claim counts and severities, i.e. 
\[\hat Y_p \;=\;\lambda_p \times \exp(\mu_p),\]
where \(\lambda_p\) and \(\mu_p\) are computed from Equation \ref{eq:frequency} and \ref{eq:severity} respectively. This gives the Bayes‐optimal expected loss and any remaining error reflects only inherent randomness in claims.
\end{enumerate}
All experiments use the synthetic data with 2\,000 policies generated as described in Section~\ref{sec:pricing}. Among them, 1\,000 policies are for training and the other 1\,000 policies are held out
for evaluation.  Each policy carries six tabular features and a
1024\,$\times$\,1024 overhead roof image.

\subsection{Predictive Performance Evaluation: Normalized Gini}
We measure predictive performance using \emph{normalized Gini coefficient}, a standard practice and widely used metric in the insurance domain to evaluate predictive models~\cite{casinstitute2019model,Ye2018CombiningPO, Pijl2017InsuranceClaims, ClaimPredictionChallenge}.  It is a rank‐based metric that captures how well predicted scores prioritize higher value observations and is appropriate for loss outcomes, which are often heavy-tailed.

Let $\{(y_i,\hat y_i)\}_{i=1}^n$ be the true responses and model predictions.  Sort the pairs by descending predicted value, yielding $\bigl(y_{(1)},\hat y_{(1)}\bigr),\dots,\bigl(y_{(n)},\hat y_{(n)}\bigr)$.  Define the cumulative true sum
\[
  C_k \;=\;\sum_{i=1}^k y_{(i)}, 
  \quad
  Y \;=\;\sum_{i=1}^n y_i.
\]
The \emph{raw} Gini coefficient is then
\[
  G_{\mathrm{raw}}(y,\hat y)
  \;=\;\frac{1}{n}\sum_{k=1}^n\frac{C_k}{Y}
        \;-\;\frac{n+1}{2n}\,.
\]
To make this metric in $[-1,1]$ and comparable across datasets, it is normalized by the “perfect” Gini achieved when $\hat y_i=y_i$:
\[
  G_{\mathrm{norm}}(y,\hat y)
  \;=\;\frac{G_{\mathrm{raw}}(y,\hat y)}{G_{\mathrm{raw}}(y,y)}.
\]
$G_{\mathrm{norm}}=1$ indicates a perfect ranking, and $G_{\mathrm{norm}}=0$ correspond to a random ordering.  $G_{\mathrm{norm}}<0$ would mean predictions are worse than random. The higher normalized Gini signals better model performance.
\subsection{Performance Gap: From Generic Pipeline to Oracle}
In Table~\ref{tab:c2_results}, we compare predictive performance across the modeling approaches described above. This highlights the gap between generic agentic AI pipelines that use only tabular data and methods used by human data scientists that incorporate domain knowledge from images. For methods using image features, the “Corr.” column shows how well the extracted variable aligns with the true underlying roof health.

The first row, \textbf{Agentic AI (Generic pipeline)}, reflects typical agentic AI workflows that use only tabular data and ignore image and domain-specific information. This approach represents the performance of a standard pipeline LLMs would generate without domain insight, achieving a normalized Gini of 0.3823. In this case, when important information is hidden in images, standard pipelines struggle to achieve good results.

The next group, \textbf{Data Scientists (Image use)}, includes several practical strategies for incorporating image information, just as a human data scientist might try. Using naive clustering of CLIP embeddings as categorical features provides some improvement (Gini 0.5042), but does not fully capture the signal (correlation with true roof health is 0.40). Feeding the full CLIP features into the model yields much better results (Gini 0.7719). Extracting the RoofHealth label from images with a vision-language model (\texttt{gpt-4o-mini}) also boosts performance (Gini 0.7271), with a much higher correlation to the true latent variable (0.81). When the model is given the true RoofHealth label as if a human labels the images perfectly, the performance almost matches the best possible (Gini 0.8310).
The clear trend is that methods using image-based domain knowledge achieve much higher predictive performance.

The last row, \textbf{Oracle (Best achievable)} represents the optimal achievable performance (Gini 0.8379), where predictions utilize the exact underlying generative mechanism and the true \texttt{RoofHealth} labels. This tier's result reflects only inherent randomness in claims data and sets a practical upper bound for predictive performance.

The improvements observed across these levels clearly demonstrate the importance of domain-specific knowledge in data science and highlight the limitations of generic, tabular-only approaches typically employed by current agentic AI.
\begin{table}[htbp]
\centering
\caption{Normalized Gini for different modeling approaches.
Rows are grouped by method type:
(1) \textbf{Agentic AI (Generic pipeline):} random forest using only tabular data;
(2) \textbf{Data Scientists (Image Use):} random forest leveraging image features or roof-health labels derived from images, simulating varying levels of domain knowledge;
(3) \textbf{Oracle (Best achievable):} model with access to the true latent variable and the generative process.
Correlation (\textbf{Corr.}) indicates how well the extracted variable aligns with the true underlying roof health.
RF = random forest; CLIP = image feature extractor.}\label{tab:c2_results}
\scalebox{0.85}{
\begin{tabular}{@{}lcc@{}}
\toprule
\textbf{Method} & \textbf{Corr.} & \textbf{Normalized Gini} \\
\midrule

\multicolumn{3}{l}{\textbf{Agentic AI (Generic pipeline)}} \\
RF (tabular only)  
    & —     & 0.3823 \\
\midrule
\multicolumn{3}{l}{\textbf{Data Scientists (Image use)}} \\
RF + CLIP clustered as 3 labels  
    & 0.4009 & 0.5042 \\
RF + CLIP features of images
    & —      & 0.7719 \\
RF + RoofHealth extracted by \texttt{gpt-4o-mini}
    & 0.8062 & 0.7271 \\
RF + true RoofHealth  
    & 1.0000  & 0.8310 \\
    \midrule
    \multicolumn{3}{l}{\textbf{Oracle (Best achievable)}} \\
Oracle (Bayes-optimal expected loss)  
    & 1.0000    & 0.8379 \\

\bottomrule
\end{tabular}
}
\end{table}

\section{Conclusion}
In this work, we illustrate that agentic AI cannot match the performance of human data scientists in our controlled setting, using a carefully designed synthetic dataset. The dataset is constructed so that an important latent variable is hidden within the image data. As a result, generic algorithms that rely solely on tabular data become insufficient, which is precisely the approach typically employed by agentic AI. In contrast, a human data scientist equipped with domain knowledge can correctly identify and utilize this latent information from the images, resulting in substantially improved performance. This underscores a limitation of current agentic AI for data science: they typically generate generic algorithms without adequately incorporating domain-specific insights. We hope this work will inspire further research into building agentic AI that can critically incorporate and utilize domain-specific knowledge, thereby bridging the gap between automated workflows and expert human performance.

\balance
\bibliographystyle{IEEEtran} 
\bibliography{refs}
\end{document}